%% file: main.tex
\documentclass[conference]{IEEEtran}
\IEEEoverridecommandlockouts

\usepackage{cite}
\usepackage{amsmath,amssymb,amsfonts}
\usepackage{graphicx}
\usepackage{booktabs}
\usepackage{multirow}
\usepackage{makecell}
\usepackage{xcolor}
\usepackage{pgfplots}
\pgfplotsset{compat=1.17}
\usepgfplotslibrary{groupplots}
\usepackage[hidelinks]{hyperref}
\usepackage{iftex}
\ifXeTeX
  \usepackage{fontspec}
\fi
\newcommand{\chat}{\ensuremath{\hat{c}}}
\newcommand{\Fmag}{\ensuremath{\lVert F\rVert}}

\begin{document}
\bstctlcite{IEEEexample:BSTcontrol}

\title{How Force Enters Matters: FiLM Conditioning\\for Behavior-Cloned VLA Policies}

\author{\IEEEauthorblockN{Chanho Lee}
\IEEEauthorblockA{LG CNS\\
Email: 2015170731@korea.ac.kr}}

\maketitle

\begin{abstract}
\input{sections/0_abstract}
\end{abstract}


\input{sections/1_intro}
\input{sections/2_related}
\input{sections/3_task}
\input{sections/4_method}
\input{sections/5_offline}
\input{sections/6_robot}
\input{sections/7_discussion}
\input{sections/8_conclusion}

\bibliographystyle{IEEEtran}
\bibliography{refs}

\end{document}

%% file: sections/0_abstract.tex
Vision-language-action (VLA) policies increasingly receive wrist wrench and contact signals in
their state, yet a behavior-cloned policy given the full wrench can still fail to bind its
behavior to it. On a real-robot suction picking task with a force-sensitive payload---a hard
contact-force limit, so that force must govern when the descent ends---a naively fine-tuned
VLA shows the familiar failure: the wrench in its state does not come to govern its behavior,
and it presses on through contact until an external safety limit intervenes. We present a conditioning pathway that changes how force is delivered, not what is
delivered: a condition vector \chat{} (contact, force magnitude, seal) computed from signals the
policy already receives and fed to the VLA through a FiLM layer. The pathway has three
properties. It is \emph{drop-in}: no backbone change, no new information, no cost in imitation
error, and, since it touches only the state input, it applies to any VLA that consumes a
proprioceptive state. It is
\emph{retrofittable}: attached to an already-trained naive policy and trained further on the
same demonstrations, it rewires how that policy responds to force. And it \emph{generalizes
beyond the demonstrations}: because force magnitude reaches the action through FiLM's affine
modulation, the policy brakes
harder the harder it is pressed, stopping and retreating at forces beyond the demonstrations,
where imitation alone leaves the response unspecified; and because the stop is tied to force
rather than to appearance, picks succeed at a stack height absent from training. We also ran it on our robot: every
FiLM-policy pick stops on its own at low contact force, whereas the naive policy drives contact
force past the limit in every attempt, and on-robot counterfactual probes reproduce the offline
crossover.

%% file: sections/1_intro.tex
\section{Introduction}
\label{sec:intro}

Behavior-cloned vision-language-action (VLA) policies are rapidly acquiring physical condition
signals. Wrist force/torque (the 6-D \emph{wrench}), contact events, and task-state bits now
routinely appear in the
observation vector, and a growing line of work fuses force into the policy backbone for
contact-rich manipulation~\cite{forcevla,phaforce,forceflow,filic}. Giving a policy the
signal, however, does not make the policy act on it. Imitation learning asks the policy to
reproduce \emph{what} the expert did, not \emph{why}: a clone can imitate the behavior
convincingly while being driven by entirely different cues than the expert was, and the loss
cannot tell such clones apart~\cite{dehaan2019causal,geirhos2020shortcut}. This paper is
about \emph{how} the force signal is delivered to the policy---a structural choice, made where
force enters the network, that decides whether the trained policy binds its behavior to force.

The consequence is easiest to see where the task hinges on force. We
study real-robot suction case picking with a single head-mounted camera and no wrist camera.
The cases carry battery cells that tolerate at most $15$\,N of contact force, so the arm must
stop descending the moment the suction cup meets the case, while stack heights differing by a
few millimeters---the entire success margin---are visually near-indistinguishable at
head-camera range. The wrench is in the state the policy receives, and every demonstration
exhibits the stop. Yet a naively fine-tuned VLA---full wrench in its state, hereafter the
\emph{naive policy}---presses through contact in every evaluation trial and never initiates a
stop of its own: each run ends only when an external safety layer, separate from the policy,
trips at the $15$\,N limit with the policy still commanding descent. The task turns on a
single transition---stop when force rises---and it is precisely this transition that the
policy fails to bind to force.

Our fix changes how force is delivered, not what is delivered. A condition vector \chat{}
(contact, normal force, seal) is computed from signals the naive policy already receives
and injected through FiLM modulation~\cite{perez2018film}; the raw wrench is masked from the
state so that the same signal does not enter twice, making \chat{} the \emph{only} pathway from
force to action. We call the policy trained behind it the \emph{FiLM policy} (``FiLM (ours)''
in tables and figures). By
construction it receives no information the naive policy lacks, and because the pathway is
explicit and unique, forcing \chat{} at inference directly measures its \emph{causal
authority}---the fraction of commanded descent the condition can cancel.

The resulting policy is not uniformly better---it is different exactly where the task is
decided. In a matched comparison---same data, same initialization, imitation error
indistinguishable---the naive policy \emph{does} respond to force at the demonstrated
magnitudes, as strongly as the FiLM policy or more so. But its response is a
\emph{template}: sized to the demonstrated forces, it fades or saturates as force rises past
them, never strengthening into a stop. The FiLM policy's response instead grows with
force, to the point of stopping and
backing away at forces beyond anything in the training data (Sec.~\ref{sec:offline}). The
mechanism carries to the real closed loop: while the naive policy drives contact force past the
$15$\,N safety limit on every attempt, the FiLM policy stops on its own in every
pick---across three stack heights, at a median contact force of $2.5$\,N---and live
counterfactual probes reproduce on the physical system the crossover seen offline, the two
responses trading places as force rises past the demonstrated range (Sec.~\ref{sec:robot}).

Our contribution is the pathway and three properties of it, each measured by intervention on
the trained policy (Sec.~\ref{sec:method-probes}):
\begin{itemize}
\item \textbf{Drop-in.} \chat{} is computed from signals the policy already receives and
  enters through an identity-initialized FiLM layer at the state token; the backbone is
  untouched, no information is added, imitation error is unchanged, and the same pathway
  attaches to another backbone ($\pi_0$), where it again turns a policy that does not stop in
  the press simulation into one that does (Sec.~\ref{sec:method},~\ref{sec:offline-pi0}).
\item \textbf{Retrofittable.} Added to an already-trained naive policy and continued on the
  same demonstrations, the pathway turns that policy's saturating force template into a
  monotone brake (Sec.~\ref{sec:offline}).
\item \textbf{Generalizes beyond the demonstrations.} Because force magnitude enters through
  FiLM's affine modulation, braking grows with force past anything demonstrated, stopping and retreating where
  imitation alone leaves the response unspecified (Sec.~\ref{sec:offline}). The same holds on
  our robot, including at a stack height absent from the demonstrations, and live
  counterfactuals reproduce the crossover (Sec.~\ref{sec:robot}).
\end{itemize}

\noindent In one sentence: \emph{from the same demonstrations, backbone, and force signal one
can obtain a policy that stops when it feels what the demonstrations felt, or one that stops
harder the harder it is pressed; the two are indistinguishable in imitation error and similar
at the demonstrated forces, and which one is obtained is decided by the pathway through which
force enters the policy.}

%% file: sections/2_related.tex
\section{Related Work}
\label{sec:related}

\textbf{Force- and tactile-conditioned VLA policies.}
ForceVLA fuses wrist force as dedicated tokens through a mixture-of-experts
backbone~\cite{forcevla}; TA-VLA injects torque adapters into the action
decoder~\cite{tavla}; PhaForce schedules policy phases on contact and force
signals~\cite{phaforce}; ForceFlow~\cite{forceflow}, FARM~\cite{farm}, and
FILIC~\cite{filic} condition diffusion or impedance-loop policies on force and tactile
input; TacForeSight learns force-conditioned tactile prediction~\cite{tacforesight}.
Each commits to a way of getting force into the network---new tokens, new experts, new
adapters, a new decoder---and each is validated by task success and retraining ablations,
sometimes alongside rollout force statistics~\cite{phaforce,forceflow}. Our pathway is the
lightest of these choices: a computed condition modulating the state token of an otherwise
unchanged backbone, attachable after training. We also evaluate differently: rather than
comparing differently trained policies by success, we intervene on the force input of a
trained policy and read the action, which exposes the \emph{form} of the response---whether
it grows or fades with force---that success rates average over. The closest such analyses
stop short of the force input: TA-VLA perturbs input tokens of its baseline~\cite{tavla},
Tactile-VLA's dose-response intervenes on the instruction~\cite{tactilevla}, and ForceFlow
masks its trained policy's visual input~\cite{forceflow}.

\textbf{Intermediate condition representations.}
Injecting a compact, task-level representation between observation and action is a recurring
design: RT-Affordance conditions on predicted affordance plans~\cite{rtaffordance}, and FiLM
modulation~\cite{perez2018film} is the standard mechanism for conditioning a backbone on side
information. Our \chat{} shares the shape but differs in role: it is computed from signals the
policy already receives rather than learned, so it adds no information and its meaning is fixed
before training.

\textbf{Causal confusion and shortcut learning.}
That imitators latch onto spurious correlates is well established: causal confusion in
IL~\cite{dehaan2019causal}, copycat behavior on observation histories~\cite{wen2020copycat},
and shortcut learning in deep networks broadly~\cite{geirhos2020shortcut,hermann2023foundations}.
Recent work documents shortcuts in generalist robot policies~\cite{shortcutgeneralist} and uses
targeted interventions to expose or correct them~\cite{interventionil2025,interventionil2023}.
The symptoms are on record in force-FiLM policies themselves: FuSe observes fine-tuned
generalists ignoring newly added sensors~\cite{fuse}, and FM-VLA diagnoses a history-length
shortcut in its own force stream and patches it with augmentation---verifying the fix, again,
only by task success~\cite{fmvla}.

\textbf{Privileged supervision and interventional evaluation.}
Privileged-teacher distillation~\cite{chen2020lbc,privilegedaction} assumes access to a signal
worth distilling; our probes instead ask how a deployed policy responds to the signal it already
has. The closest interventional precedent is ForceSight, which ablates predicted force
\emph{goals} at its controller under matched initial conditions~\cite{forcesight}; the
intervention sits on the output side of a model that takes no force input, and reads out
end-of-episode success rather than the action. Our press-retreat demonstrations are a data-side
decorrelation in the spirit of fixes for causal confusion~\cite{dehaan2019causal}, motivated
by a physical argument (Sec.~\ref{sec:discussion}).

%% file: sections/3_task.tex
\section{Task, Setting, and the Failure}
\label{sec:task}

\begin{figure}[t]
\centering
\begin{minipage}[b]{0.29\linewidth}\centering
\includegraphics[width=\linewidth]{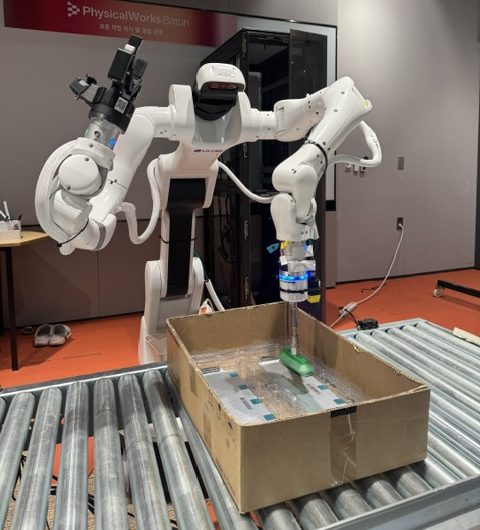}\\[-0.2em]{\scriptsize (a)}
\end{minipage}\hfill
\begin{minipage}[b]{0.42\linewidth}\centering
\input{figs/setup_pictogram}\\[0.15em]{\scriptsize (b)}
\end{minipage}\hfill
\begin{minipage}[b]{0.25\linewidth}\centering
\includegraphics[width=\linewidth]{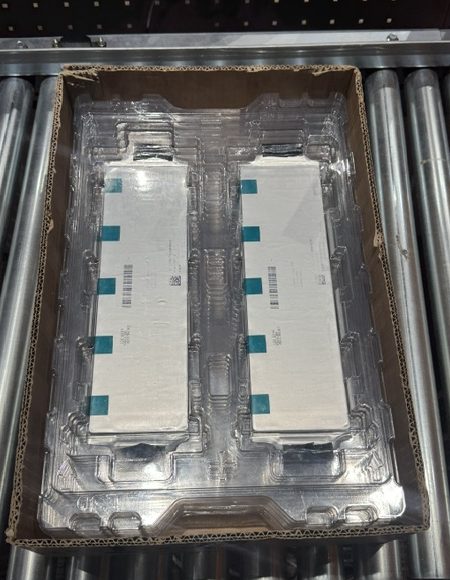}\\[-0.2em]{\scriptsize (c)}
\end{minipage}
\caption{Setting. (a)~The dual-arm mobile manipulator picking a battery-cell case from a
tray with a suction cup. (b)~The stack height varies across episodes, so the contact plane
moves by centimeters, while the head camera keeps the same pose; a few millimeters of
over-travel take contact force past the $15$\,N limit. (c)~The head-camera view of a tray:
from this range the remaining stack height is visually near-indistinguishable across layers,
so the stop-on-contact transition must be carried by force, not vision.}
\label{fig:setup}
\end{figure}

\textbf{Task and platform.}
We study suction case picking on a Dexmate Vega-1p, a dual-arm mobile manipulator: the arm
descends onto a stack of rigid cases, must stop when the suction cup meets the top case, hold
until vacuum seal is confirmed, and lift (Fig.~\ref{fig:setup}a). The cases carry battery
cells, which tolerate little compression---over-pressing risks the product, not just the
pick---so deployment enforces a $15$\,N contact-force limit. The margin is tight in both
directions: too little press and the seal fails; a few millimeters too much and contact force
climbs past the limit, since the stack is stiff (${\sim}5$\,N/mm, Sec.~\ref{sec:discussion}). The policy
controls a 7-D end-effector action ($3$-D position and $3$-D rotation increments plus a
suction command) at $15$\,Hz, executed in chunks of $5$ actions (${\sim}330$\,ms between
observations).

\textbf{Observations.}
The policy observes a single head-mounted camera (RGB and colorized depth; no wrist camera) and a
15-D state: end-effector position and quaternion, suction command, vacuum-seal bit, and the 6-D
wrist wrench. The setting is a deliberately clean diagnostic: at head-camera range, case-stack
heights differing by millimeters---the entire success margin---are visually near-indistinguishable (Fig.~\ref{fig:setup}b,c), while the wrench and seal bit signal the contact transition
directly. Every condition the task depends on is \emph{in the state the policy receives}; the
question is whether the policy binds its behavior to it.

\textbf{Demonstrations.}
We collect $100$ training episodes ($19{,}445$ frames) and hold out $6$ validation episodes
($1{,}169$ frames) with a scripted expert designed to \emph{decorrelate} force from its habitual
correlates: each episode descends with suction on, presses to a randomized force target drawn
from $[8,15]$\,N, retreats $5$--$10$\,mm from the contact plane, then hovers and lifts
immediately on seal. Episodes are collected at two different stack heights, so the contact
plane itself moves across the dataset. Every episode thus demonstrates the counterfactual the task requires:
force rises $\rightarrow$ stop $\rightarrow$ move back up, \emph{before} any seal event, at a
force level that varies independently of depth. The rationale for this design is given in
Sec.~\ref{sec:discussion}; here it matters that the naive policy and
all FiLM variants are trained on exactly these episodes.

\textbf{The failure.}
Deployed on the robot, the naive policy---a VLA fine-tuned on these demonstrations, with the
full wrench and seal bit in its state (backbone in Sec.~\ref{sec:offline})---fails every evaluation trial
the same way: it descends, makes contact, and keeps pressing---never initiating a stop of its
own---until the external safety layer aborts the run ($15.4$--$18.7$\,N measured at abort).
The transition that the demonstrations exhibit in every single episode, driven by the one signal
that observably announces it, is the transition the policy does not perform. Sec.~\ref{sec:method}
describes the pathway that changes this and the probes we use to measure the change.

%% file: figs/setup_pictogram.tex
\begin{tikzpicture}[x=1mm, y=1mm, line cap=round, line join=round]
\definecolor{pgmuted}{HTML}{767676}
\definecolor{pgcase}{HTML}{9DB4C8}
\definecolor{pgcond}{HTML}{0072B2}
\definecolor{pgnaive}{HTML}{D55E00}
\useasboundingbox (-0.3,-4) rectangle (36.4,31);
\foreach \x/\n/\lab in {0/5/{5 cases}, 21/1/{1 case}} {
  \draw[pgmuted!80, line width=0.5pt] (\x,0) -- (\x+15,0) (\x,0) -- (\x,12) (\x+15,0) -- (\x+15,12);
  \foreach \i in {1,...,\n} {
    \fill[pgcase!45] (\x+0.9,{(\i-1)*2.0+0.5}) rectangle (\x+14.1,{\i*2.0});
    \draw[pgcase!90!black, line width=0.3pt] (\x+0.9,{(\i-1)*2.0+0.5}) rectangle (\x+14.1,{\i*2.0});
  }
  \node[font=\tiny, pgmuted, anchor=north] at (\x+7.5,-0.6) {\lab};
}
\foreach \x/\ytop in {0/10.0, 21/2.0} {
  \draw[pgmuted, line width=0.6pt] (\x+7.5,20) -- (\x+7.5,{\ytop+2.6});
  \fill[pgmuted!60] (\x+6.2,{\ytop+2.6}) -- (\x+8.8,{\ytop+2.6}) -- (\x+8.2,{\ytop+1.0}) -- (\x+6.8,{\ytop+1.0}) -- cycle;
  \draw[->, pgcond, line width=0.5pt] (\x+11,{\ytop+7}) -- (\x+11,{\ytop+1.6});
}
\draw[pgmuted, densely dashed, line width=0.3pt] (-1,10) -- (16.5,10);
\draw[pgmuted, densely dashed, line width=0.3pt] (19.5,2) -- (37,2);
\draw[<->, pgmuted, line width=0.35pt] (18,10) -- (18,2);
\node[font=\tiny, pgmuted, anchor=west, inner sep=0.5pt] at (18.4,6.3) {cm};
\node[font=\tiny, pgnaive, anchor=west, inner sep=0.5pt] at (12.2,15.5) {$F<15$\,N};
\node[font=\tiny, pgcond, anchor=east, inner sep=0.5pt] at (10.2,15.5) {descend};
\fill[pgmuted!85] (15.5,26) rectangle (20.5,28.6);
\draw[pgmuted!85, line width=0.4pt] (16.8,26) -- (19.2,26) -- (18,24.8) -- cycle;
\node[font=\tiny, pgmuted, anchor=south] at (18,28.8) {head camera (fixed)};
\draw[pgmuted!55, line width=0.3pt] (18,24.8) -- (-0.5,0.3);
\draw[pgmuted!55, line width=0.3pt] (18,24.8) -- (36.5,0.3);
\end{tikzpicture}

%% file: sections/4_method.tex
\section{The Conditioning Pathway}
\label{sec:method}

The pathway has two parts: a conditioning route built so that force reaches the action in one
explicit, calibrated form (\ref{sec:method-chat}--\ref{sec:method-mask}), and a battery of
counterfactual probes that measure how much that route moves the action
(\ref{sec:method-probes}). The route is designed to be \emph{drop-in} (no backbone change, no new
information) and \emph{retrofittable} (attachable to a trained policy); it is also
\emph{measurable}: because it is explicit and unique, every component makes one number
well-defined---how much does the condition causally move the action?

\begin{figure}[t]
\centering
\includegraphics[width=\linewidth]{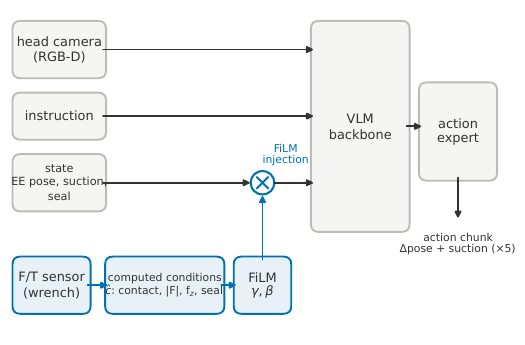}
\caption{The FiLM policy. The raw wrench does not enter the policy state; force reaches
the action only as the computed condition vector \chat{}, injected through FiLM modulation of
the state token entering the VLM backbone. Because this pathway is explicit and unique, forcing
\chat{} at inference measures the causal authority of the condition.}
\label{fig:arch}
\end{figure}

\subsection{Computed condition vector \chat{}}
\label{sec:method-chat}
From the wrench $F$ and seal bit already present in the naive policy's state we compute a small
set of task-critical conditions---the physical cues a human operator references (``if you feel
contact, stop; if the seal is on, lift''):
\begin{align}
\chat_{\mathrm{contact}} &= \mathrm{clip}\!\left(\left(\Fmag - F_0\right)/\tau,\ 0,\ 1\right),\nonumber\\
\chat_{\mathrm{fmag}} &= \left(\Fmag - F_0\right)/\tau_m \quad\text{(unsaturated)},\nonumber\\
\chat_{f_z} &= \left(f_z - o_z\right)/\tau_z,\qquad
\chat_{\mathrm{seal}} = \mathrm{seal},
\label{eq:chat}
\end{align}
with offsets and scales calibrated to the measured force statistics of the demonstrations (e.g.,
$F_0$ at the contact-onset magnitude).\footnote{Calibration quality matters and is itself a
finding. The FiLM policies in this paper are separately trained instances of the same
design: the robot picks (Sec.~\ref{sec:robot}) and the replication pair quoted in
Sec.~\ref{sec:offline-quintuple} use an earlier calibration; the ablations and the live
counterfactuals (Sec.~\ref{sec:robot-live}) use offsets re-estimated from the measured contact
distribution plus the unsaturated $\chat_{\mathrm{fmag}}$ channel.} Two properties are essential. \emph{Computed, not
learned:} \chat{} is a fixed function of observed signals, so its physical meaning cannot drift
during training---which is what makes forcing \chat{} at inference a well-defined intervention
rather than a perturbation of an arbitrary latent. \emph{Zero new information:} every channel is a deterministic
function of state dimensions the naive policy also receives, so conditioning changes
\emph{routing}, never \emph{content}---the ``it works because it got extra sensing'' objection is
excluded by construction. The channels are deliberately plural: keeping contact, force magnitude,
and seal separate is what later lets the probes attribute authority to specific signals
(Sec.~\ref{sec:offline-posthoc}).

\subsection{FiLM injection and the force mask}
\label{sec:method-mask}
\chat{} enters the policy through FiLM modulation~\cite{perez2018film} at the state token
entering the VLM backbone; the injection site matters, and Sec.~\ref{sec:offline-pi0} compares
it against the action token.
The FiLM layer is zero-initialized so that it starts as the identity, and the backbone is
untouched: at the moment the pathway is attached, the FiLM policy \emph{is} the
already-trained policy, and can be fine-tuned in place from there. Every offline FiLM
policy in this paper is obtained this way, by continuing the trained naive checkpoint
(Sec.~\ref{sec:offline}).
The policy still receives the force information---only the structure of its delivery changes.
Because \chat{} is computed \emph{from} the wrench, leaving the raw wrench in the state would
hand the policy the same signal twice, once in each form; we therefore mask the wrench
dimensions from the state (Fig.~\ref{fig:arch}). Forcing \chat{} at inference then manipulates
\emph{all} of the force information the policy receives: if the policy responds to force at
all, it must show in how the action changes with \chat{}---the quantity the probes below
measure.

\subsection{Evaluation: counterfactual probe battery}
\label{sec:method-probes}
All probes intervene on a trained, frozen policy; none involves retraining. The offline probes
(P1--P4) run on held-out validation episodes. We define
\textbf{authority} as the intervention-induced change in the commanded vertical action on
committed-descent frames, expressed as a percentage of the mean commanded descent: $0\%$ means
the intervention is ignored, $100\%$ means it fully cancels descent, ${>}100\%$ means reversal
(retreat).

\textbf{(P1) Condition forcing (open loop).} How much does the condition alone move the
action? With images and state frozen, \chat{} is forced from its observed value to a target
pattern---either all channels saturated (\emph{full forcing}) or a calibrated contact-onset
pattern matching measured first-contact statistics (\emph{contact-moment forcing}). Applicable
to FiLM models only.

\textbf{(P2) State transplant (open loop).} Does the policy respond to a \emph{real}
contact state, and through which route? The six wrench dimensions of a measured first-contact
state (the median wrench at the first force rise, pooled over episodes; the seal bit keeps its
pre-contact value) are transplanted into pre-contact descent frames. This probe applies to any
policy, including the naive policy---it is what puts both policies on the same scale. Routing
the transplant selectively, to the \chat{} computation alone ($d_{\mathrm{FiLM}}$) or the raw
state alone ($d_{\mathrm{raw}}$), attributes the response to each route.

\textbf{(P3) Dose-response sweep.} Does the policy brake \emph{more} the harder it is
pressed? A policy that has bound braking to force should; one that has memorized a contact
template should respond most near the template and fade beyond it. The transplanted wrench is
rescaled to $8$, $10$, and $12$\,N and the response traced as a function of force \emph{dose};
since the demonstrations peak near the low end of this range, the upper doses measure
extrapolation. The sweep's \emph{slope sign} is the paper's central discriminator.

\textbf{(P4) Closed-loop press simulation.} Does the policy actually stop before the force
limit when no seal ever comes? Open-loop probes cannot show consequences, so we place the
frozen policy in a closed loop at each episode's nominal first-contact timing: commanded
descent advances a virtual penetration depth, a linear-stiffness force model raises the
simulated wrench accordingly, and the seal event is withheld---the operationally critical
misalignment scenario in which no press depth ever produces a seal. Outcomes: whether the
policy stops before the force limit, and maximum penetration. A control variant grants the seal
at a shallow depth, separating force-driven stopping from seal-event-driven stopping.

\textbf{(P5) Live counterfactual probe.} Do the offline responses hold on the real robot,
where the sensor, the contact, and the control loop are real rather than replayed or
simulated? We freeze observations at staged poses
along the approach and apply (P1)--(P3) interventions to \emph{two} policies---the FiLM
policy and the naive policy---on identical frozen observations, so both receive the same
physical counterfactual.

\textbf{Scope.} P1--P3 are single-frame and open-loop; P4 closes the loop under a force model;
P5 closes the gap to the real system. No single probe is decisive---the findings below rest on
their agreement.

%% file: sections/5_offline.tex
\section{Experiments}
\label{sec:offline}

\begin{table*}[!t]
\caption{Offline battery. Upper rows: the naive policy vs.\ the FiLM policy. Lower rows: two ablation controls with the same initialization and data: the raw
wrench kept in the state (no mask), and \chat{} shuffled across the batch. Entries are the change in commanded descent caused by the intervention;
percentages are relative to the policy's own descent on the evaluated frames (\% own; none for
P1, whose frames differ). Press sim = the seal-never press simulation. The naive policy has no
\chat{} to force (---).}
\label{tab:probes}
\centering\setlength{\tabcolsep}{4.5pt}
\begin{tabular}{llrrrr}
\toprule
Policy & Force access & \multicolumn{1}{c}{\makecell{Forcing (P1)\\{\scriptsize mm/frame}}} & \multicolumn{1}{c}{\makecell{Transplant (P2)\\{\scriptsize mm/frame (\% own)}}} & \multicolumn{1}{c}{\makecell{Dose-response (P3), $8\rightarrow12$\,N\\{\scriptsize mm/frame (\% own)}}} & \multicolumn{1}{c}{\makecell{Press sim (P4)\\{\scriptsize max depth (mm)}}} \\
\midrule
naive VLA   & raw wrench               & ---                     & $+1.63$ (105\%)            & $+1.41\rightarrow+1.12$ (36$\rightarrow$28\%) $\downarrow$ & 14.9 \\
FiLM (ours) & \chat{} only              & $\mathbf{+0.9}$         & $+1.57$ (\phantom{1}94\%)  & $+1.20\rightarrow\mathbf{+4.40}$ (31$\rightarrow$\textbf{115\%}) $\uparrow$ & \textbf{9.8} \\
\midrule
FiLM, wrench kept            & raw + dead \chat{}  & $\phantom{+}0.0$ & $+1.68$ (111\%)           & $+1.57\rightarrow+1.12$ (42$\rightarrow$30\%) $\downarrow$ & 11.9 \\
FiLM, mask, shuffled \chat{} & shuffled \chat{} only & $-0.2$           & $-0.06$ (\phantom{10}0\%) & $-0.06\rightarrow-0.11$ ($\approx$0\%) $\phantom{\downarrow}$ & \textbf{57.2} \\
\bottomrule
\end{tabular}
\end{table*}

The backbone throughout is SmolVLA~\cite{smolvla}, a compact open VLA chosen so that a matched
ablation---four policies from one initialization---and the continued-training runs are
affordable; Sec.~\ref{sec:offline-pi0} repeats the central result on the $3$B-class $\pi_0$.
The offline experiments compare four policies that start from the \emph{same} trained naive
checkpoint and continue training on the \emph{same} demonstrations: the naive policy itself,
the FiLM policy, and two ablation controls:
one keeps the raw wrench in the state alongside \chat{} (no mask), the other keeps the mask but
trains on \chat{} values shuffled across the batch, so \chat{} no longer tracks the measured
force (Sec.~\ref{sec:offline-quintuple}). Imitation error on
held-out episodes is indistinguishable across all four ($0.84$--$0.96$\,mm per step): on the
data manifold, these are the same policy. Every reported model is its validation-loss-selected
(\emph{best}) checkpoint of a $20$k-step continued-training run---one uniform selection rule
for all policies.

\subsection{The two policies under intervention}
\label{sec:offline-main}

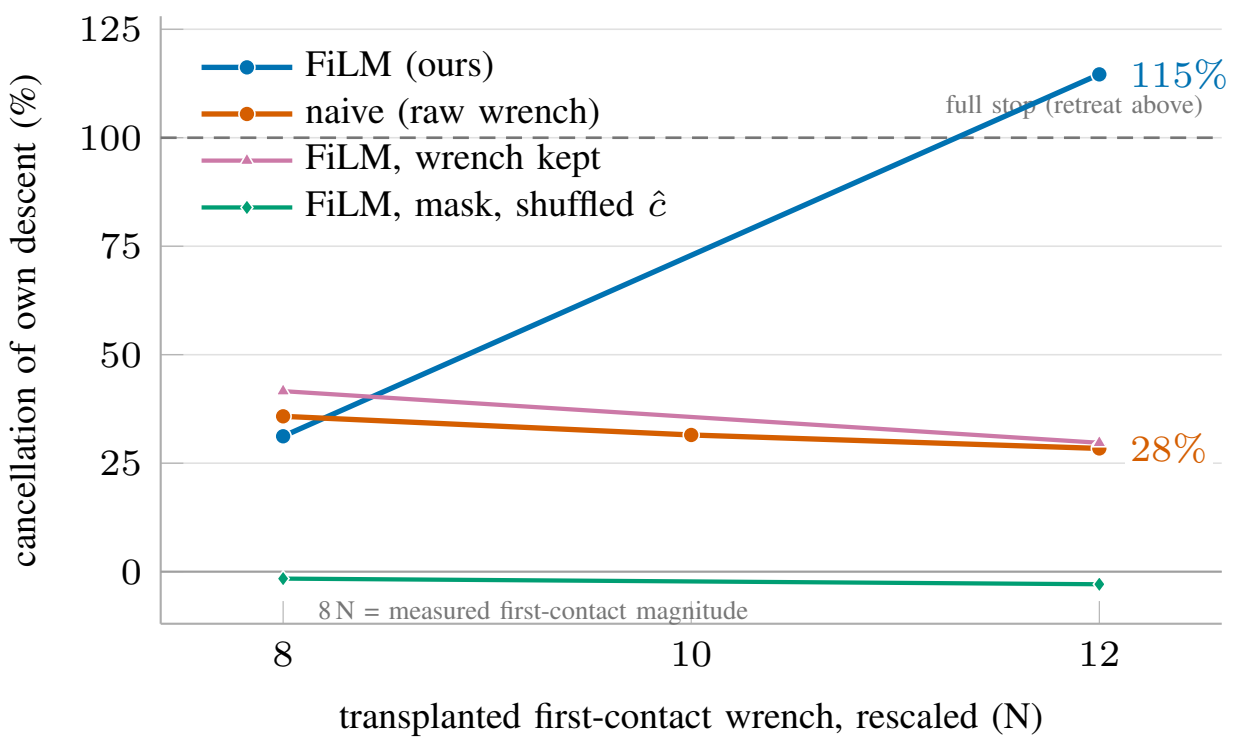
\begin{figure}[t]
\centering
\input{figs/dose_response}
\caption{Dose-response under forced wrench magnitude (P3): cancellation of each policy's
own commanded descent as the transplanted first-contact wrench is rescaled to
$8/10/12$\,N. The naive policy's response fades past the demonstrated range; the unmasked
variant lies on top of it (all response through the raw route); the shuffled-\chat{} control is
flat at zero; only the FiLM policy grows with dose, crossing full cancellation into
commanded retreat at $12$\,N.}
\label{fig:doseresponse}
\end{figure}

Table~\ref{tab:probes} (upper rows) and Fig.~\ref{fig:doseresponse} give P1--P4 for the naive
and FiLM policies; the controls follow in Sec.~\ref{sec:offline-quintuple}.

\textbf{(P1) Condition forcing.} Forcing every channel of \chat{} to saturation moves the
FiLM policy's committed descent by $+0.9$\,mm per frame. Restricting the forcing to a
calibrated contact-onset pattern shows the limit of what any policy adopts: even the
best-FiLM policy moves by only $7\%$---the 1--2-frame contact \emph{transient} is
weakly adopted across the board.

\textbf{(P2) State transplant: both policies on one scale.}
\label{sec:offline-posthoc}
Transplanting a measured first-contact state into pre-contact descent frames moves the naive
policy by $+1.63$\,mm---$105\%$ of its descent command. The naive policy is not force-blind:
its raw route responds strongly to force-bearing state. In the FiLM policy the same
transplant acts entirely through \chat{} ($+1.57$\,mm, $94\%$): changing the wrench entries of
the state while holding \chat{} fixed moves the action by $0.00$\,mm---the mask leaves no
leak. \emph{Which} state is transplanted matters as
much as whether one is: a \emph{sealed} state (seal bit on, press wrench) brakes both policies
far more than a first-contact wrench does---in the on-robot probes of Sec.~\ref{sec:robot-live},
$147\%$ vs.\ $67\%$ of own descent for the naive policy and $58\%$ vs.\ $32\%$ for the
FiLM policy. Together with
the weak contact-onset forcing above, the pattern is that stable, persistent conditions (seal,
sustained press) are adopted essentially for free, while a transient that is redundant with
depth-like cues in ordinary demonstrations is not; Sec.~\ref{sec:discussion} draws the
consequence for what mechanism can stop a stiff press.

\textbf{(P3) Dose-response: raw access binds a template, the FiLM pathway binds a brake.}
The naive policy responds to transplanted force---at the demonstrated magnitude. As the dose
rises past the demonstrations, its response \emph{fades} ($36\%\rightarrow28\%$): a template,
matched ever more poorly as reality departs from it. The FiLM policy's response
\emph{grows} with dose, crossing full cancellation into commanded retreat at $12$\,N
($31\%\rightarrow115\%$)---where the unsaturated channel reads five times its training range.
The monotone extrapolation is not learned from data (no demonstration presses at these
forces); it is inherited from the pathway: an unsaturated force scalar applied through affine
modulation. The forms
are not artifacts of checkpoint selection: no naive checkpoint we examined exceeds $65\%$
cancellation at any dose, while the FiLM policy's brake remains monotone to the end of training.
That the naive policy responds \emph{more} at the demonstrated dose does not help it: over-press
is the regime in which force rises past the template, so its brake weakens exactly as it becomes
necessary, and its strongest response---to the sealed state (Sec.~\ref{sec:robot-live})---is
bound to a signal that never arrives in an over-press.

\textbf{(P4) Press simulation: consequences.} Closing the loop under the seal-never press
turns the response forms into outcomes. Every FiLM-policy episode settles by $9.8$\,mm of
penetration, with contact force at or below $7.1$\,N when the simulation ends. The naive
policy settles in four of six episodes but runs away in two---still descending when the step
budget ends, at up to $14.9$\,mm with force climbing through $13.1$\,N; its later checkpoints
respond \emph{more} in open loop yet fare \emph{worse} here ($3/6$, max $16.0$\,mm), because a
response that saturates below full cancellation cannot stop a stiff press regardless of its
size. The seal-granted control variant localizes the failure: given a seal event at shallow
depth the naive policy stops normally ($0.5$\,mm over-travel)---withhold the seal and it
presses through. Its stopping is real but bound to the \emph{seal event}; the FiLM
policy's is bound to \emph{force}.

\subsection{Ablations: the mask and the \chat{} values}
\label{sec:offline-quintuple}

The two controls run the same battery (Table~\ref{tab:probes}, lower rows); each ablates one
element of the pathway.

\textbf{Remove the mask, and force reroutes: bypass.} The unmasked control keeps the raw
wrench in the state alongside \chat{}, and behaves as the naive policy everywhere: its
dose-response curve lies on top of the naive policy's ($+1.57\rightarrow+1.12$ vs.\
$+1.41\rightarrow+1.12$; identical at $12$\,N), forcing its \chat{} moves nothing ($0.0$\,mm),
and the pathway decomposition places its transplant response entirely on the raw route
($d_{\mathrm{FiLM}}{=}0.04$). With a redundant raw route available, gradient descent satisfies
the imitation objective through the already-functional path and never routes force through the
new one; we call this \emph{bypass}. Nothing on the demonstrations distinguishes the two
routes---the unmasked variant's imitation error matches the FiLM policy's
(Sec.~\ref{sec:offline})---so there is no pressure to move force into the pathway, and in the
retrofit setting the raw route is already trained when the identity-initialized FiLM layer is
attached. The same happens when both routes are learned from the base model: an independently
trained pair under the earlier calibration gives $7\%$ condition-forcing authority with the
wrench kept versus $76\%$ with it masked.

\textbf{Shuffle the \chat{} values, and the pathway is worse than raw access.} The shuffled control
has the same added parameters, the same continued training, the same pathway---and zero
response at every dose ($-0.2$\,mm under forcing; flat dose-response), with the worst
closed-loop outcome in the study: no episode settles, all six still descending at the step
budget, $19$--$23$\,N past the force limit, up to $57$\,mm deep. The values, not the added capacity,
carry the effect: the mask removed the raw route, and the shuffled \chat{}
replaced it with nothing.

\subsection{The pathway attaches to another backbone}
\label{sec:offline-pi0}

The same pathway---same channels, same calibration, same mask---attaches to
$\pi_0$~\cite{pi0} ($3$B-class, ${\sim}8\times$ the parameter count) at its dense state token, with the
backbone otherwise unchanged. Fine-tuned naively on the same demonstrations, $\pi_0$
reproduces the naive failure: in the seal-never press no episode settles (mean penetration
$17.0$\,mm, max $19.6$). With the pathway attached, five of six settle (mean $7.2$\,mm) at equal
imitation error ($0.93$ vs.\ $1.03$\,mm per step), and the transplant response again runs
entirely through the FiLM route (masked raw route $+0.00$\,mm). Attaching the same pathway at
the \emph{action} token instead stops no episode ($22.9$\,mm mean): the injection site matters.\footnote{This $\pi_0$ instance carries the saturating contact
channel but not the unsaturated $\chat_{\mathrm{fmag}}$, so its response is threshold-like
rather than dose-monotone, and one episode runs to $18.3$\,mm. Offline only; $\pi_0$ was not
deployed on the robot.}

%% file: figs/dose_response.tex
\definecolor{figcond}{HTML}{0072B2}
\definecolor{fignaive}{HTML}{D55E00}
\definecolor{figkept}{HTML}{CC79A7}
\definecolor{figshuf}{HTML}{009E73}
\definecolor{figmuted}{HTML}{767676}
\begin{tikzpicture}
\begin{axis}[
  width=8.6cm, height=5.6cm, xmin=7.4, xmax=12.6, xtick={8,10,12},
  ymin=-12, ymax=128, ytick={0,25,50,75,100,125},
  xlabel={transplanted first-contact wrench, rescaled (N)},
  ylabel={cancellation of own descent (\%)},
  tick label style={font=\scriptsize}, label style={font=\scriptsize},
  axis x line*=bottom, axis y line*=left,
  axis line style={figmuted!60, line width=0.4pt},
  tick style={figmuted!60},
  ymajorgrids, grid style={figmuted!22, line width=0.3pt},
  clip=false, legend style={draw=none, fill=none, font=\scriptsize,
    legend cell align=left, at={(0.02,0.98)}, anchor=north west, row sep=-1.5pt},
]
\addplot[figmuted, dash pattern=on 3pt off 2pt, line width=0.5pt, forget plot] coordinates {(7.4,100) (12.6,100)};
\node[font=\tiny, figmuted, anchor=south east] at (axis cs:12.6,100) {full stop (retreat above)};
\addplot[figmuted!70, line width=0.4pt, forget plot] coordinates {(7.4,0) (12.6,0)};
\node[font=\tiny, figmuted, anchor=west] at (axis cs:8.08,-9.5) {8\,N = measured first-contact magnitude};
\addplot[figcond, solid, line width=1.0pt, mark=*, mark size=1.5pt, mark options={solid, fill=figcond, draw=white, line width=0.3pt}] coordinates {(8,31.2) (12,114.6)};
\addlegendentry{FiLM (ours)}
\node[font=\scriptsize, figcond, anchor=west, fill=white, inner sep=1pt] at (axis cs:12.12,114.6) {$115\%$};
\addplot[fignaive, solid, line width=1.0pt, mark=*, mark size=1.5pt, mark options={solid, fill=fignaive, draw=white, line width=0.3pt}] coordinates {(8,35.8) (10,31.5) (12,28.4)};
\addlegendentry{naive (raw wrench)}
\node[font=\scriptsize, fignaive, anchor=west, fill=white, inner sep=1pt] at (axis cs:12.12,28.4) {$28\%$};
\addplot[figkept, solid, line width=0.8pt, mark=triangle*, mark size=1.5pt, mark options={solid, fill=figkept, draw=white, line width=0.3pt}] coordinates {(8,41.6) (12,29.7)};
\addlegendentry{FiLM, wrench kept}
\addplot[figshuf, solid, line width=0.8pt, mark=diamond*, mark size=1.5pt, mark options={solid, fill=figshuf, draw=white, line width=0.3pt}] coordinates {(8,-1.6) (12,-2.9)};
\addlegendentry{FiLM, mask, shuffled $\hat{c}$}
\end{axis}
\end{tikzpicture}

%% file: sections/6_robot.tex
\subsection{Real-robot experiments: closed-loop picks}
\label{sec:robot}

Beyond the offline battery, we also ran both policies on the robot of Sec.~\ref{sec:task}. The
offline results make three predictions about the real closed loop: a template response that
fades beyond the demonstrations should let contact force run away until something outside the
policy intervenes; a monotone brake should stop on its own, at forces near the demonstrated
range; and the two responses should cross between the demonstrated magnitudes and the safety
limit. The outcome variable is the \emph{force behavior} of each attempt---task success follows
from it by definition. The deployed FiLM policy is an instance of the same design
trained from the base model; the retrofitted instance of Sec.~\ref{sec:offline} was evaluated
offline and in the live probes of Sec.~\ref{sec:robot-live}, not in closed loop.

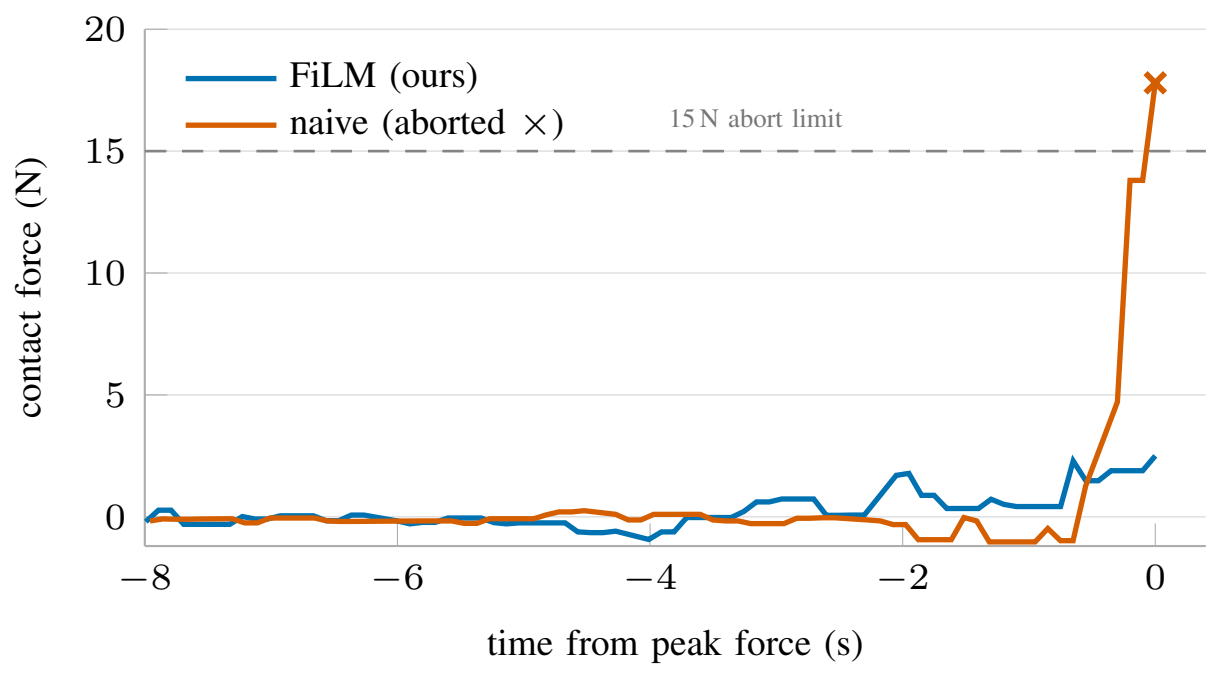
\begin{figure}[t]
\centering
\input{figs/robot_force_traces}
\caption{Contact force on the robot: one representative trial per policy, aligned at peak
force ($t{=}0$), over the interaction window (before lift, where carrying the case would
re-enter payload weight into the measured force). The naive attempt rises through the
demonstrated range and is terminated by the external $15$\,N abort ($\times$; all three naive
attempts end this way, at $15.4$--$18.7$\,N), while the FiLM-policy pick presses to
$2.5$\,N---the median of its nine picks across three stack heights, one of them absent from
the demonstrations---and stops on its own.}
\label{fig:forcetraces}
\end{figure}

The predictions hold (Fig.~\ref{fig:forcetraces}). Under the naive policy every attempt ends
the same way: contact, continued descent, and an external force-limit abort at
$15.4$--$18.7$\,N---the closed-loop realization of a response that saturates below full
cancellation. The FiLM policy stops itself in \emph{every} pick, across three stack
heights: $3/3$ at the naive policy's own height, and $3/3$ at an intermediate height absent
from the demonstrations, which contain only the other two. Peak force, measured over the
\emph{interaction window} (descent through retreat, before lifting re-enters payload weight
into the wrench; the abort monitors the same quantity), stays at or below $3.1$\,N in seven of
the nine picks (median $2.5$\,N; two picks sealed with no measurable compression at all). The
two force distributions do not overlap, and the task outcomes are their threshold readout:
$0/3$ versus $9/9$. The two harder presses ($10.2$ and $14.5$\,N) are the brake visible near
its limit---both stopped short of the abort threshold and completed the pick where the naive
policy, at the same forces, was still commanding descent.

The stopping runs through the conditioning pathway. The deployed policy's own dose-response is
monotone: transplanted wrenches at $8/10/12$\,N cancel $3\%\rightarrow7\%\rightarrow13\%$ of
committed descent---rising with dose, and carried entirely by the FiLM path (the masked raw
route contributes $+0.00$\,mm, the mask working as designed). During the picks themselves the
modulation is measurably active at the transitions ($|\gamma|$ $+48\%$, $|\beta|$ $+79\%$ at
contact/seal relative to descent), and forcing \chat{} at ten frozen poses on the robot moves
the commanded action in the correct direction at every staged condition (contact $+0.73$,
sealed $+2.09$, $+6$\,N $+2.18$\,mm/frame).

\subsection{Real-robot experiments: on-robot counterfactuals}
\label{sec:robot-live}

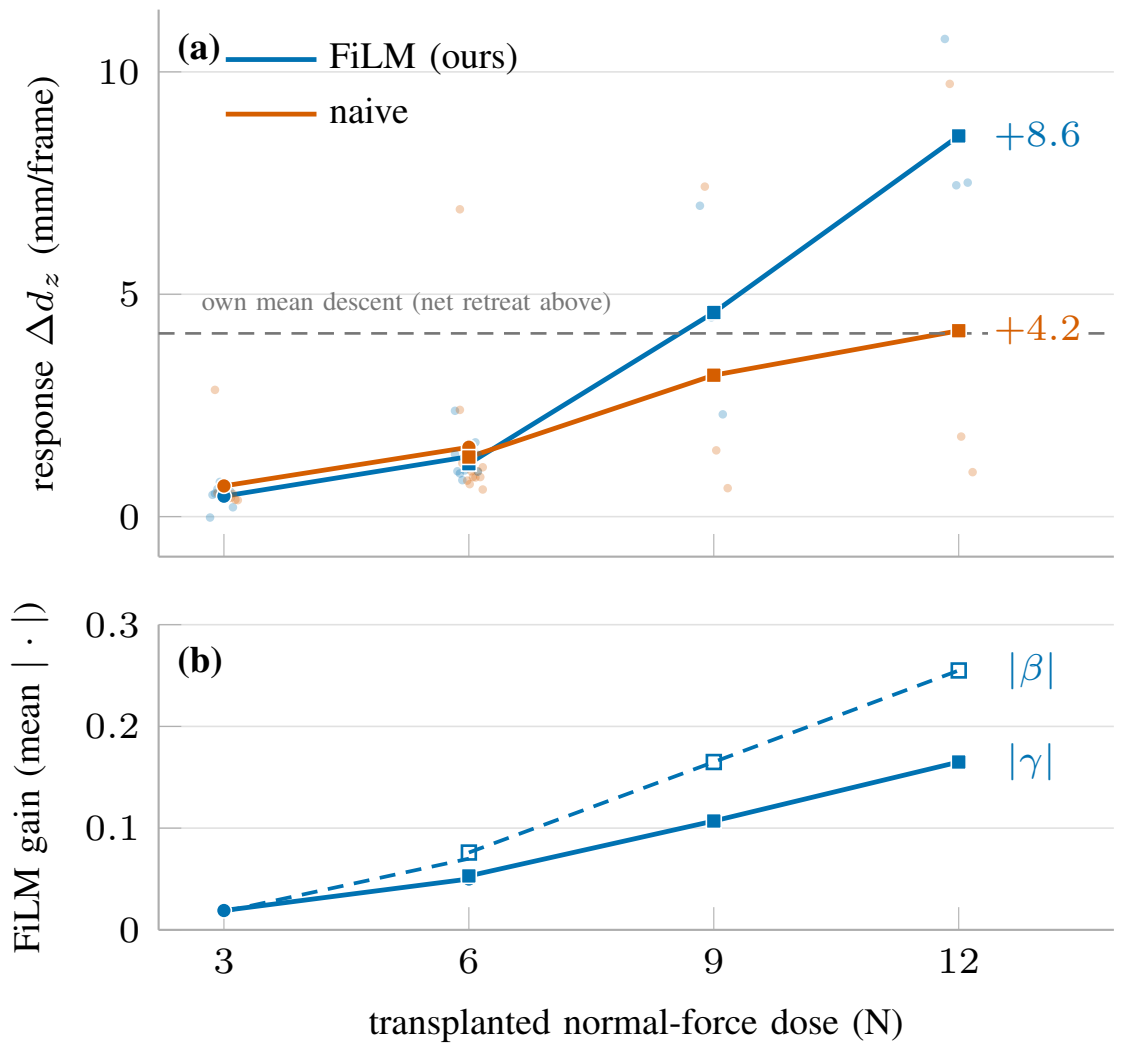
\begin{figure}[t]
\centering
\input{figs/live_counterfactuals}
\caption{On-robot counterfactuals (P5): identical frozen observations, wrench transplanted
to a fixed normal-force dose (circles: ten-pose session; squares: three-pose session; faint
dots: individual poses). (a)~The response ordering reverses between $6$ and $9$\,N---the
offline crossover on the physical system. (b)~FiLM modulation magnitude rises monotonically
with dose.}
\label{fig:live}
\end{figure}

The live probe (P5) freezes observations at staged poses and doses the naive policy and the
FiLM policy with identical counterfactual states. Two sessions matter
(Fig.~\ref{fig:live}).

\textbf{Fair dose.} At transplant doses matching the demonstrations ($\leq$\,contact
magnitude), the \emph{naive} policy responds as strongly or stronger than the FiLM
policy (first-contact: $67\%$ vs.\ $32\%$ of own descent; sealed state: $147\%$ vs.\
$58\%$)---in quantitative agreement with the offline dose-response curves at the same doses,
where the template peaks (and, as Sec.~\ref{sec:offline-main} argues, this does not help it).
The FiLM policy's response is dose-monotone across the staged conditions (hover $+0.29$
$\rightarrow$ pre-seal $+0.87$ $\rightarrow$ sealed $+1.41$ $\rightarrow$ $+6$\,N
$+1.70$\,mm/frame).

\textbf{High dose.} Raising the forced normal force to $+6/+9/+12$\,N reproduces the offline
crossover on the robot: the FiLM policy accelerates its braking, $+1.19 \rightarrow
+4.59 \rightarrow +8.56$\,mm/frame---at $12$\,N, twice its own descent command, i.e., commanded
retreat---while the naive policy's response collapses at the margin ($+1.34 \rightarrow +3.18
\rightarrow +4.18$; increments $+1.84 \rightarrow +1.00$). The reversal sits between $6$ and
$9$\,N. At near-surface poses the separation is starkest: naive $+1.0$--$1.8$ vs.\ FiLM
$+7.5$\,mm/frame at $12$\,N. Similar at the demonstrated doses, categorically different where
the task is decided---on the robot as offline.

Three validity notes. Force-torque baselines were re-anchored at a pre-descent hover before
each probe session, and absolute transplant targets were drift-shifted so both policies
received the same physical dose; live FiLM-policy magnitudes are conservative lower
bounds, as some frozen states are slightly off-manifold for its input distribution. The naive
policy's closed-loop attempts were run at a single stack height---its failure mode is uniform
and force-driven, but the height-generalization evidence applies to the FiLM policy
only. High-dose probes cover three poses; we report them as targeted supplementary
measurements, not a survey.

%% file: figs/robot_force_traces.tex
\definecolor{figcond}{HTML}{0072B2}
\definecolor{fignaive}{HTML}{D55E00}
\definecolor{figmuted}{HTML}{767676}
\begin{tikzpicture}
\begin{axis}[
  width=8.6cm, height=5.0cm, xmin=-8, xmax=0.4, ymin=-1.2, ymax=20,
  xtick={-8,-6,-4,-2,0}, ytick={0,5,10,15,20},
  xlabel={time from peak force (s)}, ylabel={contact force (N)},
  tick label style={font=\scriptsize}, label style={font=\scriptsize},
  axis x line*=bottom, axis y line*=left,
  axis line style={figmuted!60, line width=0.4pt},
  tick style={figmuted!60},
  ymajorgrids, grid style={figmuted!22, line width=0.3pt},
  clip=false, legend style={draw=none, fill=none, font=\scriptsize,
    legend cell align=left, at={(0.02,0.98)}, anchor=north west, row sep=-1.5pt},
]
\addplot[figmuted!90, dash pattern=on 4pt off 3pt, line width=0.5pt, forget plot] coordinates {(-8,15) (0.4,15)};
\node[font=\tiny, figmuted, anchor=south west] at (axis cs:-4.0,15.2) {15\,N abort limit};
\addplot[figcond, line width=1.0pt] coordinates {(-7.992,-0.20) (-7.892,0.27) (-7.792,0.27) (-7.692,-0.31) (-7.592,-0.31) (-7.328,-0.31) (-7.228,0.01) (-7.128,-0.09) (-7.028,-0.09) (-6.928,0.04) (-6.665,0.04) (-6.565,-0.16) (-6.465,-0.16) (-6.365,0.07) (-6.265,0.07) (-6.001,-0.17) (-5.901,-0.29) (-5.801,-0.23) (-5.701,-0.23) (-5.601,-0.05) (-5.337,-0.05) (-5.237,-0.24) (-5.137,-0.29) (-5.037,-0.25) (-4.937,-0.25) (-4.672,-0.25) (-4.572,-0.62) (-4.472,-0.65) (-4.372,-0.65) (-4.272,-0.59) (-4.010,-0.93) (-3.910,-0.62) (-3.810,-0.62) (-3.710,-0.03) (-3.610,-0.03) (-3.357,-0.03) (-3.257,0.22) (-3.157,0.61) (-3.057,0.61) (-2.957,0.73) (-2.704,0.73) (-2.604,0.06) (-2.504,0.06) (-2.404,0.07) (-2.304,0.07) (-2.051,1.70) (-1.951,1.78) (-1.851,0.88) (-1.751,0.88) (-1.651,0.34) (-1.400,0.34) (-1.300,0.72) (-1.200,0.51) (-1.100,0.42) (-1.000,0.42) (-0.750,0.42) (-0.650,2.29) (-0.550,1.48) (-0.450,1.48) (-0.350,1.89) (-0.100,1.89) (0.000,2.50)};
\addlegendentry{FiLM (ours)}
\addplot[fignaive, line width=1.0pt] coordinates {(-7.959,-0.17) (-7.859,-0.09) (-7.759,-0.10) (-7.659,-0.10) (-7.559,-0.09) (-7.309,-0.08) (-7.209,-0.25) (-7.109,-0.25) (-7.009,-0.05) (-6.909,-0.05) (-6.657,-0.05) (-6.557,-0.17) (-6.457,-0.19) (-6.357,-0.19) (-5.820,-0.17) (-5.571,-0.17) (-5.471,-0.27) (-5.371,-0.27) (-5.271,-0.08) (-5.171,-0.08) (-4.921,-0.08) (-4.821,0.08) (-4.721,0.20) (-4.621,0.20) (-4.521,0.25) (-4.272,0.10) (-4.172,-0.13) (-4.072,-0.13) (-3.972,0.10) (-3.872,0.10) (-3.603,0.10) (-3.503,-0.13) (-3.403,-0.17) (-3.303,-0.17) (-3.203,-0.28) (-2.940,-0.28) (-2.839,-0.06) (-2.739,-0.06) (-2.639,-0.04) (-2.539,-0.04) (-2.276,-0.13) (-2.176,-0.17) (-2.076,-0.32) (-1.976,-0.32) (-1.876,-0.94) (-1.615,-0.94) (-1.515,-0.03) (-1.414,-0.17) (-1.314,-1.03) (-1.214,-1.03) (-0.951,-1.03) (-0.851,-0.48) (-0.751,-0.98) (-0.651,-0.98) (-0.551,1.27) (-0.300,4.71) (-0.200,13.80) (-0.100,13.80) (0.000,17.80)};
\addlegendentry{naive (aborted $\times$)}
\addplot[only marks, mark=x, mark size=2.6pt, fignaive, line width=1.1pt, forget plot] coordinates {(0.000,17.80)};
\end{axis}
\end{tikzpicture}

%% file: figs/live_counterfactuals.tex
\definecolor{figcond}{HTML}{0072B2}
\definecolor{fignaive}{HTML}{D55E00}
\definecolor{figmuted}{HTML}{767676}
\begin{tikzpicture}
\begin{groupplot}[
  group style={group size=1 by 2, vertical sep=4.5mm,
    xlabels at=edge bottom, xticklabels at=edge bottom},
  width=7.9cm, xmin=2.2, xmax=13.9, xtick={3,6,9,12},
  tick label style={font=\scriptsize}, label style={font=\scriptsize},
  axis x line*=bottom, axis y line*=left,
  axis line style={figmuted!60, line width=0.4pt},
  tick style={figmuted!60},
  ymajorgrids, grid style={figmuted!22, line width=0.3pt},
  clip=false, legend style={draw=none, fill=none, font=\scriptsize,
    legend cell align=left, at={(0.05,0.98)}, anchor=north west},
]
\nextgroupplot[height=5.2cm, ymin=-0.9, ymax=11.4,
  ylabel={response $\Delta d_z$ (mm/frame)}]
\addlegendimage{figcond, line width=0.9pt}\addlegendentry{FiLM (ours)}
\addlegendimage{fignaive, line width=0.9pt}\addlegendentry{naive}
\addplot[only marks, mark=*, mark size=0.6pt, figcond, opacity=0.28, forget plot] coordinates {(2.83,-0.02) (2.86,0.49) (2.89,0.52) (2.92,0.62) (2.95,0.78) (2.99,0.40) (3.02,0.52) (3.05,0.51) (3.08,0.56) (3.11,0.21)};
\addplot[only marks, mark=*, mark size=0.6pt, figcond, opacity=0.28, forget plot] coordinates {(5.83,2.38) (5.86,1.02) (5.89,0.97) (5.92,0.82) (5.95,1.04) (5.99,1.28) (6.02,1.44) (6.05,1.53) (6.08,1.67) (6.11,1.39)};
\addplot[only marks, mark=*, mark size=0.6pt, fignaive, opacity=0.28, forget plot] coordinates {(2.89,2.85) (2.92,0.53) (2.95,0.47) (2.98,0.41) (3.01,0.58) (3.05,0.42) (3.08,0.42) (3.11,0.51) (3.14,0.38) (3.17,0.37)};
\addplot[only marks, mark=*, mark size=0.6pt, fignaive, opacity=0.28, forget plot] coordinates {(5.89,6.91) (5.92,1.20) (5.95,1.21) (5.98,0.81) (6.01,0.73) (6.05,0.89) (6.08,0.88) (6.11,1.02) (6.14,0.89) (6.17,1.11)};
\addplot[only marks, mark=*, mark size=0.6pt, figcond, opacity=0.28, forget plot] coordinates {(5.83,1.40) (5.97,1.15) (6.11,1.01)};
\addplot[only marks, mark=*, mark size=0.6pt, figcond, opacity=0.28, forget plot] coordinates {(8.83,6.99) (8.97,4.48) (9.11,2.30)};
\addplot[only marks, mark=*, mark size=0.6pt, figcond, opacity=0.28, forget plot] coordinates {(11.83,10.74) (11.97,7.45) (12.11,7.51)};
\addplot[only marks, mark=*, mark size=0.6pt, fignaive, opacity=0.28, forget plot] coordinates {(5.89,2.40) (6.03,1.02) (6.17,0.61)};
\addplot[only marks, mark=*, mark size=0.6pt, fignaive, opacity=0.28, forget plot] coordinates {(8.89,7.42) (9.03,1.49) (9.17,0.64)};
\addplot[only marks, mark=*, mark size=0.6pt, fignaive, opacity=0.28, forget plot] coordinates {(11.89,9.73) (12.03,1.80) (12.17,1.00)};
\addplot[figcond, line width=0.9pt, mark=*, mark size=1.4pt, mark options={fill=figcond, draw=white, line width=0.3pt}, forget plot] coordinates {(3.00,0.46) (6.00,1.35)};
\addplot[fignaive, line width=0.9pt, mark=*, mark size=1.4pt, mark options={fill=fignaive, draw=white, line width=0.3pt}, forget plot] coordinates {(3.00,0.69) (6.00,1.56)};
\addplot[figcond, line width=0.9pt, mark=square*, mark size=1.4pt, mark options={fill=figcond, draw=white, line width=0.3pt}, forget plot] coordinates {(6.00,1.19) (9.00,4.59) (12.00,8.56)};
\addplot[fignaive, line width=0.9pt, mark=square*, mark size=1.4pt, mark options={fill=fignaive, draw=white, line width=0.3pt}, forget plot] coordinates {(6.00,1.34) (9.00,3.18) (12.00,4.18)};
\addplot[figmuted, dash pattern=on 3pt off 2pt, line width=0.5pt, forget plot] coordinates {(2.2,4.12) (13.9,4.12)};
\node[font=\tiny, figmuted, anchor=south west] at (axis cs:2.5,4.12) {own mean descent (net retreat above)};
\node[font=\scriptsize, figcond, anchor=west, fill=white, inner sep=1pt] at (axis cs:12.35,8.56) {$+8.6$};
\node[font=\scriptsize, fignaive, anchor=west, fill=white, inner sep=1pt] at (axis cs:12.35,4.18) {$+4.2$};
\node[font=\scriptsize\bfseries, anchor=north west] at (rel axis cs:0.0,0.99) {(a)};
\nextgroupplot[height=3.6cm, ymin=0, ymax=0.30, ytick={0,0.1,0.2,0.3},
  ylabel={FiLM gain (mean $|\cdot|$)},
  xlabel={transplanted normal-force dose (N)}]
\addplot[figcond, line width=0.9pt, mark=*, mark size=1.4pt, mark options={fill=figcond, draw=white, line width=0.3pt}, forget plot] coordinates {(3.00,0.019) (6.00,0.050)};
\addplot[figcond, line width=0.7pt, dash pattern=on 3pt off 2pt, mark=, mark size=1.3pt, mark options={solid, fill=white, line width=0.5pt}, forget plot] coordinates {(3.00,0.018) (6.00,0.070)};
\addplot[figcond, line width=0.9pt, mark=square*, mark size=1.4pt, mark options={fill=figcond, draw=white, line width=0.3pt}, forget plot] coordinates {(6.00,0.053) (9.00,0.107) (12.00,0.165)};
\addplot[figcond, line width=0.7pt, dash pattern=on 3pt off 2pt, mark=square, mark size=1.3pt, mark options={solid, fill=white, line width=0.5pt}, forget plot] coordinates {(6.00,0.076) (9.00,0.165) (12.00,0.255)};
\node[font=\scriptsize, figcond, anchor=west] at (axis cs:12.35,0.165) {$|\gamma|$};
\node[font=\scriptsize, figcond, anchor=west] at (axis cs:12.35,0.255) {$|\beta|$};
\node[font=\scriptsize\bfseries, anchor=north west] at (rel axis cs:0.0,0.99) {(b)};
\end{groupplot}
\end{tikzpicture}

%% file: sections/7_discussion.tex
\section{Discussion and Limitations}
\label{sec:discussion}

\textbf{Why the demonstrations must be anticipatory.}
A first-principles calculation explains both the failure and the fix. The case stack is stiff:
${\sim}5$\,N/mm of penetration. Stopping below the $15$\,N limit therefore requires a full stop
within ${\sim}2$\,mm of contact---under $330$\,ms action chunks and the demonstrated descent
rates, this is arithmetically impossible for \emph{reactive} gating that begins braking at the
contact transient, no matter how much authority the transient carries. The viable mechanisms are
anticipatory: approach slowly, press deliberately, retreat. The press-retreat demonstrations
teach exactly this behavior prior, and the probes confirm the resulting division of labor: the
successful policy's contact-moment authority is small ($7\%$) while its full-condition authority
is large ($76\%$)---success comes from an anticipatory behavior prior plus post-hoc state
gating, not from a reflex.

\textbf{A computed condition is an operational knob.}
Because \chat{} is computed rather than learned, the pathway creates knobs the naive policy lacks. Force-torque baselines drift
(${\sim}1$\,N within a day, pose-sensitive); the FiLM policy re-anchors its offsets at a
hover pose each run---a deployment-time correction with no retraining---while the baseline's
force statistics are frozen in its normalization.

\textbf{Limitations.}
Single task, single platform, small robot-$n$: the robot results are a case study, and the
statistical claims live in the offline battery ($n{=}6$ held-out episodes $\times$ dense frames,
seven policies across two backbones). The retrofitted instance and the $\pi_0$ instance were
not deployed in closed loop, and the press-retreat demonstrations are a precondition we do not
ablate here. The press simulation depends on a linear force model;
we mitigate with the seal-granted control and the live probes, which bracket the simulated
mechanism from both sides. Contact-moment (reactive) authority plateaus low in every
configuration we trained---in development phases, oversampling transition frames also failed to
move it; distilling an explicit
descend-until-contact oracle (e.g., via DAgger~\cite{ross2011dagger,chen2020lbc}) is the natural
next step, as is a negative-control task whose critical condition is \emph{not} observable in
the state, and testing whether force-fusion architectures that keep the raw signal alongside a
dedicated route~\cite{forcevla,forceflow} also bypass it.

%% file: sections/8_conclusion.tex
\section{Conclusion}
\label{sec:conclusion}

On a picking task where force must govern when the descent ends, a behavior-cloned VLA policy
given the full wrench did not let that signal govern its behavior, pressing on until an
external safety limit intervened. We changed how force is delivered, not what is delivered: a
condition vector computed from signals the policy already receives, fed to the backbone
through a FiLM layer. The pathway is drop-in, attaches to an already-trained policy, carries to
another backbone, and---because it applies force magnitude through affine modulation---brakes
harder the harder it is pressed, stopping and retreating beyond the demonstrated forces and
succeeding at an undemonstrated stack height. Because the condition is explicit, the same
pathway can be interrogated: forcing it at inference reads out how much, and in what form, the
policy responds to force. From the same demonstrations, backbone, and signal, how force enters
the policy decided what the policy learned to do with it.

%% file: main.bbl
\begin{thebibliography}{10}
\providecommand{\url}[1]{#1}
\csname url@samestyle\endcsname
\providecommand{\newblock}{\relax}
\providecommand{\bibinfo}[2]{#2}
\providecommand{\BIBentrySTDinterwordspacing}{\spaceskip=0pt\relax}
\providecommand{\BIBentryALTinterwordstretchfactor}{4}
\providecommand{\BIBentryALTinterwordspacing}{\spaceskip=\fontdimen2\font plus
\BIBentryALTinterwordstretchfactor\fontdimen3\font minus
  \fontdimen4\font\relax}
\providecommand{\BIBforeignlanguage}[2]{{%
\expandafter\ifx\csname l@#1\endcsname\relax
\typeout{** WARNING: IEEEtran.bst: No hyphenation pattern has been}%
\typeout{** loaded for the language `#1'. Using the pattern for}%
\typeout{** the default language instead.}%
\else
\language=\csname l@#1\endcsname
\fi
#2}}
\providecommand{\BIBdecl}{\relax}
\BIBdecl

\bibitem{forcevla}
J.~Yu \emph{et~al.}, ``{ForceVLA}: Enhancing {VLA} models with a force-aware
  {MoE} for contact-rich manipulation,'' \emph{arXiv preprint
  arXiv:2505.22159}, 2025.

\bibitem{phaforce}
M.~Wang \emph{et~al.}, ``{PhaForce}: Phase-scheduled visual-force policy
  learning with slow planning and fast correction for contact-rich
  manipulation,'' \emph{arXiv preprint arXiv:2603.08342}, 2026.

\bibitem{forceflow}
S.~Zhang \emph{et~al.}, ``{ForceFlow}: Learning to feel and act via
  contact-driven flow matching,'' \emph{arXiv preprint arXiv:2605.11048}, 2026.

\bibitem{filic}
H.~Ge \emph{et~al.}, ``{FILIC}: Dual-loop force-guided imitation learning with
  impedance torque control for contact-rich manipulation tasks,'' \emph{arXiv
  preprint arXiv:2509.17053}, 2025.

\bibitem{dehaan2019causal}
P.~de~Haan, D.~Jayaraman, and S.~Levine, ``Causal confusion in imitation
  learning,'' in \emph{NeurIPS}, 2019, arXiv:1905.11979.

\bibitem{geirhos2020shortcut}
R.~Geirhos \emph{et~al.}, ``Shortcut learning in deep neural networks,''
  \emph{Nature Machine Intelligence}, vol.~2, pp. 665--673, 2020.

\bibitem{perez2018film}
E.~Perez, F.~Strub, H.~de~Vries, V.~Dumoulin, and A.~Courville, ``{FiLM}:
  Visual reasoning with a general conditioning layer,'' in \emph{AAAI}, 2018.

\bibitem{tavla}
Z.~Zhang \emph{et~al.}, ``{TA-VLA}: Elucidating the design space of
  torque-aware vision-language-action models,'' in \emph{CoRL}, 2025,
  arXiv:2509.07962.

\bibitem{farm}
E.~Helmut, N.~Funk, T.~Schneider, C.~de~Farias, and J.~Peters,
  ``Tactile-conditioned diffusion policy for force-aware robotic
  manipulation,'' \emph{arXiv preprint arXiv:2510.13324}, 2025.

\bibitem{tacforesight}
Y.~Zang \emph{et~al.}, ``{TacForeSight}: Force-guided tactile world model for
  contact-rich manipulation,'' \emph{arXiv preprint arXiv:2606.11184}, 2026.

\bibitem{tactilevla}
J.~Huang, S.~Wang, F.~Lin, Y.~Hu, C.~Wen, and Y.~Gao, ``{Tactile-VLA}:
  Unlocking vision-language-action model's physical knowledge for tactile
  generalization,'' \emph{arXiv preprint arXiv:2507.09160}, 2025.

\bibitem{rtaffordance}
S.~Nasiriany \emph{et~al.}, ``{RT-Affordance}: Affordances are versatile
  intermediate representations for robot manipulation,'' \emph{arXiv preprint
  arXiv:2411.02704}, 2024.

\bibitem{wen2020copycat}
C.~Wen, J.~Lin, T.~Darrell, D.~Jayaraman, and Y.~Gao, ``Fighting copycat agents
  in behavioral cloning from observation histories,'' in \emph{NeurIPS}, 2020,
  arXiv:2010.14876.

\bibitem{hermann2023foundations}
K.~Hermann, H.~Mobahi, T.~Fel, and M.~C. Mozer, ``On the foundations of
  shortcut learning,'' \emph{arXiv preprint arXiv:2310.16228}, 2023.

\bibitem{shortcutgeneralist}
Y.~Xing, X.~Luo, J.~Xie, L.~Gao, H.~Shen, and J.~Song, ``Shortcut learning in
  generalist robot policies: The role of dataset diversity and fragmentation,''
  \emph{arXiv preprint arXiv:2508.06426}, 2025.

\bibitem{interventionil2025}
Y.~Chen, Y.~Zhang, G.~D'urso, N.~Lawrance, and B.~Tidd, ``Improving
  generalization ability of robotic imitation learning by resolving causal
  confusion in observations,'' \emph{arXiv preprint arXiv:2507.22380}, 2025.

\bibitem{interventionil2023}
S.~Pfrommer, Y.~Bai, H.~Lee, and S.~Sojoudi, ``Initial state interventions for
  deconfounded imitation learning,'' \emph{arXiv preprint arXiv:2307.15980},
  2023.

\bibitem{fuse}
J.~Jones, O.~Mees, C.~Sferrazza, K.~Stachowicz, P.~Abbeel, and S.~Levine,
  ``Beyond sight: Finetuning generalist robot policies with heterogeneous
  sensors via language grounding,'' in \emph{ICRA}, 2025, arXiv:2501.04693.

\bibitem{fmvla}
R.~Li \emph{et~al.}, ``{FM-VLA}: Force-based memory for vision-language-action
  models in contact-rich manipulation,'' \emph{arXiv preprint
  arXiv:2607.18231}, 2026.

\bibitem{chen2020lbc}
D.~Chen, B.~Zhou, V.~Koltun, and P.~Kr{\"a}henb{\"u}hl, ``Learning by
  cheating,'' in \emph{CoRL}, 2019.

\bibitem{privilegedaction}
X.~Mao, Y.~Xu, Z.~Sun, E.~Miller, D.~Layeghi, and M.~Mistry, ``Learning
  long-horizon robot manipulation skills via privileged action,'' \emph{arXiv
  preprint arXiv:2502.15442}, 2025.

\bibitem{forcesight}
J.~A. Collins, C.~Houff, Y.~L. Tan, and C.~C. Kemp, ``{ForceSight}: Text-guided
  mobile manipulation with visual-force goals,'' in \emph{ICRA}, 2024,
  arXiv:2309.12312.

\bibitem{smolvla}
M.~Shukor \emph{et~al.}, ``{SmolVLA}: A vision-language-action model for
  affordable and efficient robotics,'' \emph{arXiv preprint arXiv:2506.01844},
  2025.

\bibitem{pi0}
K.~Black \emph{et~al.}, ``$\pi_0$: A vision-language-action flow model for
  general robot control,'' \emph{arXiv preprint arXiv:2410.24164}, 2024.

\bibitem{ross2011dagger}
S.~Ross, G.~J. Gordon, and J.~A. Bagnell, ``A reduction of imitation learning
  and structured prediction to no-regret online learning,'' in \emph{AISTATS},
  2011.

\end{thebibliography}
